\pdfoutput=1

\documentclass[conference]{IEEEtran}
\IEEEoverridecommandlockouts

\usepackage{pbalance}

\IEEEoverridecommandlockouts 
\usepackage{graphicx}
\usepackage{algorithmic}
\usepackage{algorithm}
\usepackage[numbers]{natbib}
\usepackage[OT4,T1]{fontenc}
\usepackage[cmex10]{amsmath}
\usepackage{url}
\usepackage{multirow}
\usepackage{float}

\title{Social Media, Topic Modeling and Sentiment Analysis in Municipal Decision Support}

\author{
\IEEEauthorblockN{Miloš Švaňa}
\IEEEauthorblockA{\\
VSB - Technical Univesity of Ostrava\\
Faculty of Economics\\
17. listopadu 2172/15, 708 00 Ostrava, Czechia\\
Email: milos.svana@vsb.cz}

}

\begin{document}
\maketitle

\begin{abstract}
Many cities around the world are aspiring to become \textit{smart}. However, smart initiatives often give little weight to the opinions of average citizens.

Social media are one of the most important sources of citizen opinions. This paper presents a prototype of a framework for processing social media posts with municipal decision-making in mind. The framework consists of a sequence of three steps: (1) determining the sentiment polarity of each social media post (2) identifying prevalent topics and mapping these topics to individual posts, and (3) aggregating these two pieces of information into a fuzzy number representing the overall sentiment expressed towards each topic. Optionally, the fuzzy number can be reduced into a tuple of two real numbers indicating the "amount" of positive and negative opinion expressed towards each topic.

The framework is demonstrated on tweets published from Ostrava, Czechia over a period of about two months. This application illustrates how fuzzy numbers represent sentiment in a richer way and capture the diversity of opinions expressed on social media.
\end{abstract}

\section{Introduction}
\IEEEPARstart{M}{any cities} around the world are aspiring to become \textit{smart}. A \textit{Smart City} is characterized by an extensive use of information technologies to support municipal decision-makers in effective resource utilization \citep{vinod_kumar_smart_2017}. This ICT support can be applied in many areas, including traffic and transportation, waste management, accommodation or culture.

Ideally, municipal planning involves multiple stakeholder groups: local authorities, businesses, average citizens and commuters, or environmental activists. In practice, however, decisions are mostly based on an interplay between authorities and businesses offering Smart City technologies. Opinions of average citizens are often given little weight \citep{halegoua_2020, cardullo_whose_2019}.

Some cities use surveys to gather opinions on at least some issues and projects. Although useful, surveys have many disadvantages. Most importantly, by asking only a predetermined set of questions and letting the respondent choose only from a low number of possible answers they limit the respondent's ability to fully express their thoughts and opinions. This prevents decision makers from serendipitously discovering unexpected issues and ideas.

Social media represent an alternative source of citizen opinions. They offer much greater freedom of expression as users can create new content whenever they want and instead of being limited by a set of questions, they can use free-form text, images, videos or audio. This freedom together with significant content creation velocity might also be a disadvantage. Relatively simple statistical methods for processing survey data are inadequate for social media. Instead, advanced techniques from fields such as machine learning, computer vision, or natural language processing have to be used.

This paper presents a prototype of a framework for extracting information from social media with the aim to support municipal decision-making. The framework combines topic modeling techniques with sentiment analysis. First, the system detects topics discussed on social media at a specific location. Then it evaluates sentiment towards each topic. To capture uncertainty arising from different people having different opinions on the same topic, this sentiment is modelled as a triangular fuzzy number (TFN). The TFN representation, however, might not be understood by people without background knowledge. Therefore a calculation of a "degree of conformity" with fuzzy sets representing the concepts of positive and negative opinion follows. The result can be interpreted as an "amount" of positive/negative opinion expressed towards a specific topic.

The remainder of the paper is organized as follows: In section \ref{sec:review} I review literature related to topic modeling, sentiment analysis and their applications in the context of Smart Cities. Section \ref{sec:framework} discusses the framework design. Section \ref{sec:data} then describes data used in experiments and preprocessing procedures. Demonstration of the framework application on test data can be found in section \ref{sec:demo}. Finally, section \ref{sec:discussion} discusses the limitations of the framework and future research.

\section{Literature review}
\label{sec:review}

\subsection{Topic Modeling}

Topic modeling can be understood as (1) a "statistical technique for revealing the underlying semantic structure in a large collection of documents", or (2) "a technique comes with group of algorithms that reveal, discover and annotate thematic structure in collection of documents" \citep{kherwa_topic_2018}.

In practice, topic modeling involves taking a corpora of text documents and discovering various topics discussed in these documents. Topics are usually represented by a set of most relevant terms \citep{churchill_evolution_2022}. Once topics are identified, a mapping between topics and individual documents can be created. Depending on the method used, one document can be assigned to one or multiple topics.

There is a plethora of topic modeling methods. In \citep{kherwa_topic_2018}, the authors distinguish between two categories: (1) algebraic methods, usually based on some form of word-document matrix factorization and (2) statistical methods. On the other hand, \citep{churchill_evolution_2022} provides a chronological overview of the development of topic modeling methods. Other reviews, e.g., \citep{vayansky_review_2020} then focus on selecting the best method for a given dataset or application.

One of the most popular topic modeling methods is Latent Dirichlet Allocation (LDA). This method is based on the idea that documents in a given corpora are generated by a probabilistic process. Each document can be understood as a mixture of topics, with each topic representing a probability distribution over words from some vocabulary \citep{Blei2001LatentDA}. However, the framework presented in this paper relies on a more recent method called BERTopic \citep{grootendorst2022bertopic}. It outperforms LDA on multiple benchmark datasets both in coherence and topic diversity -- two common topic modeling evaluation metrics \citep{churchill_evolution_2022}.

\subsection{Sentiment analysis}

Sentiment analysis is "an approach that uses Natural Language Processing (NLP) to extract, convert and
interpret opinion from a text and classify them into positive, negative or natural sentiment" \citep{drus_sentiment_2019}. There are two main groups of sentiment analysis methods (1) lexicon-based and (2) methods based on supervised learning models.

Lexicon-based methods use sentiment lexicons containing information about sentiment polarity of different words. Individual word polarities in a given document are looked up and then aggregated into an overall document polarity.

Methods based on supervised learning models rely on various machine learning techniques to train a classification model for predicting document sentiment. Naive Bayes is one of the most popular approaches in this category. Supervised methods require a training dataset that contains a ground truth sentiment for each document. Many product and service review websites let their users combine textual reviews with some sort of a numerical rating scale, e.g., a 5-star rating system. This data can be easily used to train a sentiment classifier.

One of the contributions of this paper lies in the application of fuzzy modelling methods in sentiment analysis. Several researchers have already explored this path. In \citep{jefferson_fuzzy_2017} authors describe a fuzzy expert system for sentiment analysis. There have also been attempts at designing a sentiment analysis method that utilizes a fuzzy thesaurus \citep{ismail_semantic_2018}. However, to our best knowledge, fuzzy methods have not been used to aggregate sentiment expressed towards multiple documents.

There have been multiple applications of sentiment analysis in the context of Smart Cities and urban planning \citep{dahbi_social_2019, li_new_2016}. There are many opportunities and challenges related to the use of sentiment analysis in urban planning, including visualization, multilingual audiovisual opinion mining, or peer-to-peer opinion mining tools for citizens \citep{Ahmed2016SentimentAF}.

\section{Framework Design}
\label{sec:framework}

\subsection{Topic Modeling}

As mentioned, the framework prototype uses a topic modeling method called BERTopic. Originally proposed in \citep{grootendorst2022bertopic}, BERTopic solves the problem of topic modeling by combining word embeddings with hierarchical clustering. The procedure consists of the following steps:

\begin{enumerate}
    \item \textbf{Create an embedding for each document}. One of the downsides of traditional methods such as LDA is that they represent documents in an bag-of-words fashion. This representation ignores both order of words in a document and their semantic relationships. BERTopic relies on an embedding representation. Embeddings are vectors that are able to somewhat capture the semantics of words or documents. Documents with similar meaning should be represented by similar vectors. As the name suggests, BERTopic uses embeddings based on BERT \citep{devlin2019bert}.
    \item \textbf{Reduce embedding dimensionality}. Embeddings can have hundreds or even thousands of dimensions. When it comes to clustering, high dimensionality causes multiple issues. \citep{grootendorst2022bertopic}. First, the difference between the distance of the nearest point to a cluster centre and the distance between the furthest point from a cluster centre shrinks \citep{aggarwal_dimensions}. Second, lower number of dimensions leads to better performance both in terms of time and clustering accuracy. BERTopic therefore uses UMAP \citep{mcinnes2020umap} to reduce the number of dimensions.
    \item \textbf{Use a clustering algorithm to create document clusters}. BERTopic uses HDBSCAN with document embeddings reduced by UMAP as input. HDBSCAN is a hierarchical algorithm able to create a tree of cluster-subcluster structures. BERTopic user can set the number of clusters/topics to be generated by the algorithm. To provide the desired number of clusters, small similar topics are merged together.
    \item \textbf{Create topic representations}. Similarly to other topic modeling methods, BERTopic represents each topic as a list of words. To find words best describing a given topic BERTopic uses a modified TF-IDF score \citep{grootendorst2022bertopic} calculated as:
    \begin{equation}
        W_{t,c} = tf_{t,c}\ \log(1 + \frac{A}{tf_t})
    \end{equation}
    where $tf_{t,c}$ represents the frequency of term $t$ in cluster $c$. $A$ is the average number of words per cluster and $tf_t$ is the total frequency of term $t$ across all clusters.
\end{enumerate}

\subsection{Sentiment Analysis}

The TextBlob\footnote{https://github.com/sloria/TextBlob} Python library was selected for the framework prototype. This choice was made mainly for pragmatic reasons - ease of installation, use and integration with other parts of the framework.

TextBlob offers two sentiment analysis methods: a lexicon and pattern-based one and a pretrained Naive Bayes model. In the former, TextBlob uses a polarity lexicon and structural patterns to determine both the degree of polarity and the degree of subjectivity. The Naive Bayes model was trained to classify movie reviews as positive or negative. Instead of providing degrees of polarity and subjectivity as output, this method returns the probabilities of a given text being positive and negative. Being the default, the lexicon and pattern-based method was also used in presented experiments. The output of this model carries a certain degree of uncertainty which can be exploited when aggregating sentiment analysis results with fuzzy methods.

\subsection{Fuzzy Aggregation}

Once sentiment polarity and topics are extracted from a set of social media posts, both pieces of information are combined to provide an overall view of what is being discussed in a given municipality and whether the population perceives a given topic positively or negatively.

Simple aggregation metrics such as arithmetic mean would lead to a loss of information. For example, there might be a controversial topic with many positive, but also some negative opinions. Arithmetic mean might present the topic sentiment as slightly positive. When presented with this information, the user cannot tell whether the aggregated opinion is slightly positive because a majority of individual opinions is slightly positive or because there are many positive opinions counterbalanced by a small number of negative opinions.

To address this issue, topic sentiment is modeled as a triangular fuzzy number (TFN). Assuming we know both the sentiment polarity of each social media post, and their topic distribution, the TFN core can be determined as a weighted mean with topic distribution percentages serving as weights. For instance, consider 3 topics with polarities and topic distributions depicted in table \ref{tab:aggr_example}. The core of the TFN representing the sentiment towards topic 1 can be calculated as:
\begin{equation}
    m_{t1} = \frac{0.5 \cdot 0.5 + 0.35 \cdot 0.3 - 0.2 \cdot 0.2}{0.5 + 0.3 + 0.2} = 0.315
\end{equation}
Weighted standard deviation is then used to determine the TFN support interval. As in the case of determining the TFN core, the degree to which a given post belongs to a given topic should determine the strength of its influence on the shape of the support interval. Therefore a weighted variant of standard deviation is used, the general formula of which being:
\begin{equation}
    \sigma = \sqrt{\frac{\sum_{i=1}^N w_i(x_i - x^*)^2}{\frac{M-1}{M}\sum_{i=1}^N w_i}}
\end{equation}
with $N$ representing the total sample size, $x^*$ representing weighted mean, and $M$ the number of non-zero weights.

Once the weighed standard deviation is known, the TFN support interval is calculated as:
\begin{equation}
\label{eq:support}
    [m_{t1} - s\sigma; m_{t1} + s\sigma]
\end{equation}
where $s$ is a positive real number that determines the scaling between the length of the support interval and weighted standard deviation. Its influence on the aggregation result is a subject of further research.

\begin{table}[!ht]
\centering
\caption{Example of an input for the aggregation procedure}
\label{tab:aggr_example}
\begin{tabular}{|r|r|rrr|}
\hline
\multicolumn{1}{|l|}{} & \multicolumn{1}{l|}{\multirow{2}{*}{\textbf{Polarity}}} & \multicolumn{3}{c|}{\textbf{Topic distribution}}                                                                                  \\ \cline{3-5} 
                       & \multicolumn{1}{l|}{}                                   & \multicolumn{1}{r|}{\textbf{Topic 1}} & \multicolumn{1}{r|}{\textbf{Topic 2}} & \textbf{Topic 3} \\ \hline
\textbf{Post 1}        & 0.5                                                     & \multicolumn{1}{r|}{0.5}                         & \multicolumn{1}{r|}{0.3}                         & 0.2                         \\ \hline
\textbf{Post 2}        & 0.35                                                    & \multicolumn{1}{r|}{0.3}                         & \multicolumn{1}{r|}{0.4}                         & 0.3                         \\ \hline
\textbf{Post 3}        & -0.2                                                    & \multicolumn{1}{r|}{0.2}                         & \multicolumn{1}{r|}{0.2}                         & 0.6                         \\ \hline
\end{tabular}

\end{table}

Taking inspiration from \citep{zapletal_three-level_2023}, the next step of the aggregation procedure is the calculation of a degree of conformity between the TFN describing topic sentiment and fuzzy sets representing the concepts of a positive and negative opinion. The result of this step is a tuple of two numbers from interval $[0; 1]$ that could be interpreted as the overall "amount" of positive and negative opinion expressed towards a given topic. Although there is some loss of information when compared to the original TFN, the two numbers are more likely to be understood by a layperson with no knowledge of fuzzy set theory.

The fuzzy metric of possibility \citep{klir1995fuzzy} is used to calculate the degree of conformity, e.g.:
\begin{equation}
\label{eq:opinion_amount}
Pos(\tilde{A},\tilde{PO}) = \sup_{x \in X} \min\big(\mu_{\tilde{A}}(x),\mu_{\tilde{PO}}(x)\big)
\end{equation}
$A$ denotes aggregated topic sentiment with $\mu_{\tilde{A}}(x)$ being its membership function. Similarly $PO$ represents the concept of a positive opinion with a membership function $\mu_{\tilde{PO}}(x)$.

\section{Data and Preprocessing}
\label{sec:data}

Social media posts published on Twitter were used to demonstrate the proposed framework. This social network  was chosen specifically because it lets researchers easily access its data through an API. Using this API, a dataset of approximately 3000 tweets published by users located in Ostrava, Czechia was created. These tweets cover the period of January and February of 2023.

Given the location, most tweets in the dataset are written in Czech. There are, however, several tweets written in other languages, such as English, Slovak or Polish. To address the issue of multilingualism, as well as the fact that TextBlob uses an English lexicon when analyzing sentiment, the DeepL Translator\footnote{https://www.deepl.com/translator} API was used to automatically translate each post to English.

Several preprocessing steps were then applied on translated tweets:
\begin{itemize}
    \item removal of tweets that are shorter than 60 characters under the assumption that it might be difficult to reliably extract sentiment and topics from short tweets,
    \item removal of non-letter characters,
    \item removal of hashtags, usernames and URLs,
    \item turning all text into lowercase,
    \item removal of stop words.
\end{itemize}

\section{Framework Demonstration}
\label{sec:demo}

The framework prototype was implemented as a Jupyter notebook. To summarize, after the input data was preprocessed, the following sequence of steps was applied:
\begin{enumerate}
    \item TextBlob was used to determine the sentiment polarity of each post. Polarity of each post is represented as a real number from a $[-1, 1]$ interval with -1 representing absolutely negative polarity and positive number representing absolutely positive polarity.
    \item BERTopic was used to identify topics and create a mapping between topics and individual tweets. Provided output follows the format in table \ref{tab:aggr_example}.
    \item A set of TFNs representing aggregated topic sentiment was constructed. The scaling constant $s$ in equation \ref{eq:support} was set to 1.
    \item Degrees of conformity between each topic's TFN and the concepts "positive opinion" and "negative opinion" were determined. 
\end{enumerate}

Overall, BERTopic identified 42 different topics. Topics with the highest overall "mass" calculated as the sum of the topic's probability across all tweets are listed in table \ref{tab:topics}.

\begin{table*}[h!]
\centering
\caption{Most prevalent topics identified by BERTopic in tweets published from Ostrava.}
\label{tab:topics}
\begin{tabular}{|r|l|r|r|r|}
\hline
\textbf{No.} &
\textbf{Topic}                                                                        & \textbf{Prevalence} & \textbf{Positivity} & \textbf{Negativity} \\ \hline
1 & 'vote', 'election', 'politics', 'party', 'ano', 'government', 'people'                & 41.98               & 0.997               & 0.190               \\ \hline
2 &'czech', 'republic', 'czech republic', 'czechs', 'czk', 'inflation'                   & 33.68               & 0.810               & 0.434               \\ \hline
3 &'boy', 'date', 'girl', 'crying', 'apologize', 'shes', 'ive'                           & 27.19               & 0.812               & 0.390               \\ \hline
4 &'school', 'education', 'teacher', 'class', 'educated', 'academics', 'students'        & 25.93               & 0.897               & 0.287               \\ \hline
5 & 'game', 'field', 'leader', 'go', 'games', 'team', 'match'                             & 25.89               & 0.839               & 0.390               \\ \hline
6 & 'area', 'city', 'building', 'ostrava', 'construction', 'also', 'mappa'                & 25.32               & 0.745               & 0.473               \\ \hline
7 & 'hydrogen', 'gt', 'car', 'anymore', 'hydrogen bike', 'product', 'hydrogen technology' & 19.47               & 0.741               & 0.459               \\ \hline
8 &'something nothing', 'number', 'nothing', 'something', 'old figure', 'number thumbs'  & 18.81               & 0.760               & 0.272               \\ \hline
9 & 'miracles', 'ribbons', 'expect miracles', 'political responsibility', 'imagine'       & 18.72               & 0.757               & 0.344               \\ \hline
\end{tabular}
\end{table*}

Presented topic distribution intuitively makes sense. During given time period, one of the most dominant topics in public discourse was the upcoming presidential election which corresponds to the first topic in table \ref{tab:topics}. One can also see that not all topics are necessarily relevant to municipal decision making, for example topic no. 3 in the table. At the same time, topics such as no. 9 are difficult to interpret.

\begin{figure*}[tbp]
    \centering
    \label{fig:tfns}
    \includegraphics[width=0.6\textwidth]{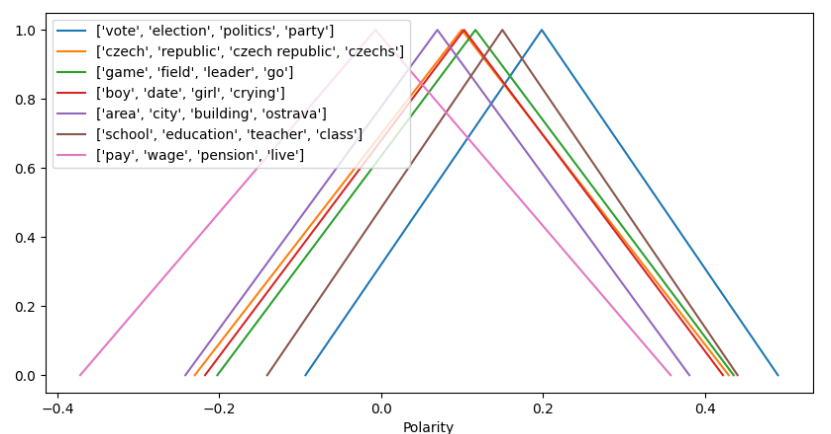}
    \caption{Depiction of TFNs representing overall sentiment towards selected topics extracted from tweets published from Ostrava}.
\end{figure*}

TFNs representing the several aggregated topic sentiments are depicted in figure \ref{fig:tfns}. It can be deduced that the topic \textit{vote, election, politics, party} is perceived quite positively and compared to topic \textit{pay, wage, pension, live} it also has a narrower support. This indicates a lower level of opinion diversity. The latter topic is also perceived most negatively, at least among the topics displayed in the figure.

Table \ref{tab:topics} also contains information about the conformity with the concepts of positive and negative opinion. The membership function of the fuzzy set representing positive opinion has a value of 0 until polarity value 0, then grows linearly to 1 until polarity reaches 0.2 and then has a value of 1. The "negative opinion" fuzzy set is its mirror image. 

It can be seen that the information provided by these numbers tells a story similar to the TFN visualization. For example, when comparing topics \textit{school, education, teacher, class} and \textit{area, city, building, ostrava}, one could conclude that the former is not only perceived more positively, but that it is also less controversial given the lower "amount" of negative opinion.

\section{Discussion}
\label{sec:discussion}

As demonstrated, the framework prototype can be used to create a representation of opinions expressed towards different topics on social media posted by citizens of a specific municipality. The prototype, however, employs certain simplifications that should be addressed in future research. 

The TFNs representing topic sentiments are currently symmetric. This is might not be the best reflection of reality. Ways of making the TFNs asymmetric should be explored. Metrics such as weighted skewness or semivariance could be used. Next, Twitter users have an option to "like" or "retweet" a post created by someone else. These actions can represent approval and could be therefore used as aggregation weights. Finally, other topic-modeling and sentiment-analysis methods should be tested and compared to the existing TextBlob+BERTopic stack. Moreover, TextBlob provides additional information beyond sentiment polarity: the subjectivity degree of each document. This metric could again be used as weight.

Framework output can be used directly by municipal decision makers to make more informed decisions. However, there are other possible applications. One of them is comparison of different municipalities. It should be possible to compare the TFNs of a specific topic across multiple cities. If additional information such as municipal budgets is available, one could also deploy methods such as Data Envelopment Analysis to evaluate their efficiency. Topic sentiments could be also aggregated into an overall sentiment expressed towards everything happening in a given municipality.

However, using the proposed framework as a replacement for other methods of gathering citizen opinion might not be the best course of action. Instead, the framework should play a complementary role, as social media users might not accurately represent the overall population. Groups such as the elderly might be underrepresented. At the same time, methods such as surveys can provide biased information too. Combining the proposed framework with surveys might lead to a better overall representation of citizen opinion than either of these methods separately.

\section{Acknowledgment}
This paper was supported by the SGS project No. SP2023/078. This support is gratefully acknowledged.

\bibliography{fedcsis}

\end{document}